\documentclass[10pt,twocolumn,letterpaper]{article}

\usepackage{iccv}
\usepackage{times}
\usepackage{epsfig}
\usepackage{graphicx}
\usepackage{amsmath}
\usepackage{amssymb}

\usepackage[pagebackref=true,breaklinks=true,letterpaper=true,colorlinks,bookmarks=false]{hyperref}

\iccvfinalcopy 

\def\iccvPaperID{3211} 

\ificcvfinal\fi
\begin{document}

\title{Structured Light Phase Measuring Profilometry Pattern Design for Binary Spatial Light Modulators}

\author{Daniel L. Lau, Yu Zhang, Kai Liu\\
University of Kentucky\\
}

\maketitle

\begin{abstract}
Structured light illumination is an active 3-D scanning technique based on projecting/capturing a set of striped patterns and measuring the warping of the patterns as they reflect off a target object's surface. In the case of phase measuring profilometry (PMP), the projected patterns are composed of a rolling sinusoidal wave, but as a set of time-multiplexed patterns, PMP requires the target surface to remain motionless or for scanning to be performed at such high rates that any movement is small. But high speed scanning places a significant burden on the projector electronics to produce contone patterns inside of short exposure intervals. Binary patterns are, therefore, of great value, but converting contone patterns into binary comes with significant risk.  As such, this paper introduces a contone-to-binary conversion algorithm for deriving binary patterns that best mimic their contone counterparts.  Experimental results will show a greater than $3\times$ reduction in pattern noise over traditional halftoning procedures.
\end{abstract}

\section{Introduction}
Structured light illumination (SLI)~\cite{REF:Geng2011SLITutorial,REF:Salvi2010SLIReview,REF:Salvi2004SLIReview} is a 3-D imaging process that digitally reconstructs target surfaces through active triangulation between a camera and a projector. SLI methods that use a single, continuously projected pattern typically employ a pseudo-random dot or stripe pattern.  For instance, the original Microsoft Kinect employed an incoherent, near-infrared, dot pattern with a wavelength-matched, near-IR camera that is able to  discern pattern dots from ambient light~\cite{REF:Khoshelham2012Kinect}. The problem for single pattern SLI systems, such as the Kinect, is that they rely on matching small sub-windows between the projected and captured images~\cite{REF:Kawasaki2013OneShot}, resulting in rough edges along discontinuities in depth (i.e. a step edge). Single-pattern SLI system can not, therefore, reconstruct thin objects. 

In order to individually resolve every pixel of the camera sensor, multi-pattern SLI procedures are available that rely on digital modulation techniques such as quadrature amplitude~\cite{REF:Mosino2010PSIQuadrature} and phase modulation~\cite{REF:Liu2011PhaseModulation}. In these instances, real-time scanning is achieved by multiplexing patterns in color with one pattern projected as red, another as green, and a third as blue~\cite{REF:Su2007ColorSLI}; however, such  schemes have practical limitations for polychromatic surfaces. Temporal multiplexing is, many times, the default method for multi-pattern SLI, but as a scanning process, this approach becomes susceptible to distortion caused by object motion~\cite{REF:Wang2011SLICoding}. In response, many investigators attempt to accelerate their hardware in the hopes that the object's observed motion remains small.  In this same vein, many researchers have attempted to minimize the number of patterns to also minimize the total amount of observed motion.

Taken hand-in-hand with high speed image acquisition are short exposures~\cite{REF:Gong2010SLIUltraFast, REF:Zhang2010SLISuperFast}, and for structured light, short exposures may create concerns associated with the spatial light modulator in the projector~\cite{REF:Chen2013SLM, REF:Wissmann2011SLM}.  For instance, DLP modulators from Texas Instruments~\cite{REF:Sun2013Imaging3D} use microelectromechanical mirrors that flip back and forth with intermediate shades of gray produced by rapidly flipping between these states. The human visual system, by blurring the projected light pulses in time, sees shades of gray, but with a short camera exposure, SLI systems may inadvertently posterize the structured patterns, introducing additional distortion on top of motion and sensor noise.

In some cases, the method used by a DLP projector to produce contone pixels can be exploited for SLI as performed by Gong~{\it et al}~\cite{REF:Gong2010SLIUltraFast} who used a high speed camera to record the changing light pattern. Short of having a 1,000 fps camera though, the posterization of patterns caused by short camera exposures has a detrimental effect that can be combated using binary SLI patterns, and in order to convert contone patterns to binary, the standard technique of digital halftoning is specifically tasked. In addressing the application of digital halftoning to PMP pattern design, Dai~{\it et al}~\cite{REF:Dai2013DitherPattern} showed that not all halftoning methods are equal. 

Of all the many halftoning procedures available, Analoui and Allebach's Direct Binary Search (DBS)~\cite{REF:AnalouiAllebachDirectBinarySearch} is, perhaps, the ideal choice since it is generally regarded as one of the best methods for minimizing the visual artifacts between the original contone image and its binary counterpart. DBS works by iteratively swapping binary pixels within small neighborhoods until the binary image, after low-pass filtering, closely mimics the original contone image.  The quality of the final halftone depends largely on the selected low-pass filter, which is intended to model the human visual system. Because of the similarities between the human visual system and the modulation transfer function of a projector lens, DBS should be as good as any halftoning scheme for mapping contone SLI patterns to binary.  The resulting patterns can be made contone by defocusing the projector lens and should then be immune to changes in exposure time.

While independently halftoning the component SLI patterns will minimize visual artifacts of the projected patterns, experimental results show that dithering leads to pattern noise in the resulting 3-D surface reconstructions, and in light of these artifacts, we propose a novel dithering technique that correlates the halftoning of each component pattern such that we minimize the effects on phase, regardless of visual integrity. In other words, our goal is to dither the pixels of the component patterns such that differences between the defocused halftone and the contone original at one pattern index are canceled out by differences at another.

Like DBS, the proposed halftoning method relies on defocusing the projector during scanning and is easy to tune to a specific amount of defocus just as DBS can be tuned to varying viewing conditions. And our algorithm maintains a substantial degree of freedom for contone pattern design to allow, for example, for dual-frequency patterns as prescribed by Liu {\it et al}~\cite{REF:Liu2010Realtime3D} or to detect motion, as prescribed by  Lau {\it et al}~\cite{REF:Lau2010MotionDetection}, and to do so without modifying the process by which patterns are processed after capture. And as will be demonstrated both theoretically and experimentally, our algorithm achieves these many goals while producing as much as a $3\times$ reduction in pattern noise versus traditional forms of dithering.

\section{Related Work}
The idea of using a binary patterns in a digital projector and then relying on projector defocus to produce contone structured light patterns is generally regarded as having benefits with relation to minimizing the effects of gamma distortion in the digital projector~\cite{Lei:09,REF:Gong2010SLIUltraFast}; however, it is also known that binary dithering techniques import an artificial texture of their own in much the same way dithering imparts a visually distressing texture in printed images~\cite{ulichney}. In the work of Li~{\it et al}~\cite{Li2014236}, the authors do an extensive study on the use of various dither algorithms for minimizing dither noise in their 3D reconstructions, looking at Bayer's dither, several different error diffusion variants; however, their results were limited to a 3-pattern SLI technique championed by Zhang~\cite{doi:10.1117/1.2402128}. The major determination made by Li~{\it et al} was that for high frequency patterns, their best binary renditions were simple square waves with 50\% duty cycles. 

Lohry and Zhang~\cite{Lohry2012917} focus on improving the 50\% duty cycle square waves and assuming that 2-by-2 bins of pixels could be treated like a contone pixel with 5 gray levels.  Their results showed a reduction to 30\% of the dither noise over the simple square waves. Zuo~{\it et al}~\cite{Zuo:12} similarly looked at improving the 50\% duty cycle square waves by employing 1-D pulse-width modulation (or dithering) to generate a 2-D pattern, something they called sinusoidal pulse width modulation (SPWM). 

In the work of Dai and Zhang~\cite{Dai2013790,Dai201479}, the authors focus on the use of a single, dithered, sinusoidal pattern that gets circular shifted, as is the case with contone patterns, but they iteratively toggled pixels in the single binary pattern that ultimately reduced the phase error once the three shifted renditions were phase processed. So the optimization was performed using a DBS-like algorithm, as is being proposed in this paper.

In the follow up study by Dai~{\it et al}~\cite{REF:Dai2013DitherPattern}, the authors look at the same algorithm as~\cite{Dai201479} but this time optimize the spatial domain appearance of the patterns, showing that the spatially optimized patterns perform better over a wider range of pattern defocus where the phase optimized pattern works best when the projector is near-focused.  Sun~{\it et al} follow a similar path in that they focus on a single binary pattern that gets circular shifted three times, but they offer an improved error metric. Like the works of~\cite{Sun2015158}, the algorithm is greatly limited in the types of patterns that it can be used, mainly a single pattern that renders a single spatial frequency and that must be shifted a multiple of 3 steps. To our knowledge, no one has introduced a flexible dithering framework for general sinusoidal gratings with arbitrary spatial frequencies over an arbitrary number of frames, and that is the goal of this paper to introduce such a framework.

\section{Phase Measuring Profilometry}
As described in Lau {\it et al}~\cite{REF:Lau2010MotionDetection}, SLI attempts to reconstruct a 3-D object's surface by triangulating between the pixels of a digital camera, with coordinates $(x_c,y_c)$, and a projector, with coordinates $(x_p,y_p)$.  Assuming that the camera is epipolar rectified to the projector along the $x$ axis, one need only match $(x_c,y_c)$ to a $y_p$ coordinate. As such, SLI attempts to encode or modulate the $y_p$ coordinates of the projector over a sequence of $N$ projected patterns such that a unique $y_p$ value can be obtained by demodulating the captured camera pixels.

In the case of phase measuring profilometry (PMP), the projected image pixels are defined by the set, $\{I_p:n=0,1,\ldots,N-1\}$, according to:
\begin{equation}
I_p[n] = \frac{1}{2} + \frac{1}{2} \cos \left( 2\pi (\frac{n}{N}-y_p)\right),
\label{EQ:ProjectorPattern}
\end{equation}
where $y_p$ is the row coordinate of the pixel in the projector in the range from 0 to 1, $I_p$ is the intensity of that pixel with dynamic range from 0 to 1, and $n$ is the phase-shift index over the $N$ total patterns. The corresponding pixel in the captured images, observed in the presence of ambient light, can then be expressed as:
\begin{equation}
    I_c[n] =  A_c + B_c \cos\left( 2\pi (\frac{n}{N}-y_p) \right) + \eta
    \label{EQ:CameraPattern}
\end{equation}
where $I_c$ is the intensity of that pixel in the $n$th captured image while $A_c$ represents the averaged amount of projector light reflected back to the camera over the $N$ patterns plus the ambient light. The term, $B_c$, represents the fraction of light, coming from the projector, that is reflected back to the camera with smaller values for darker surfaces. The term, $\eta$, derives from additive Gaussian noise in the sensor, which will be ignored in all remaining discussion.

As an $N$-point vector of scalar values, we can apply the discrete-time Fourier transform (DFT) to compartmentalize ambient and modulated light into specific DFT coefficients such that $A_c$ becomes the $k=0$ coefficient while $B_c\cos(\cdot)$ goes to the $k=1$ and $k=N-1$ complex conjugate pair.  For the purpose of 3-D reconstruction, we only need this $k=1$ coefficient to estimate $y_p$ according to the coefficient's phase value:
\begin{equation}
y_p = \angle I_c[k=1] = \arctan \left\{\frac{{\cal I}\{ I_c[k=1] \}}{{\cal R}\{ I_c[k=1] \}}\right\},
\label{EQ:Phase}
\end{equation}
while also measuring the quality of the estimate based upon its magnitude:
\begin{equation}
    B_c  = \left \Vert I_c[k=1] \right\Vert = \left\{ {I_c[k=1]}^2+{I_c[k=1]}^2 \right\}^\frac{1}{2}.
\end{equation}
Of the remaining $k=2, \ldots, N-2$ coefficients, all should be zero unless employed for carrying higher frequency phase terms, as proposed by Liu {\it et al}~\cite{REF:Liu2010Realtime3D} who used the $k=2$ and $k=N-2$ conjugate pair to produce dual-frequency PMP patterns, or to detect motion, as proposed by Lau {\it et al}~\cite{REF:Lau2010MotionDetection} who used it to delete scan points irrevocably corrupted by object motion.

\begin{figure}[t]
\centerline{\includegraphics[width=3.30in]{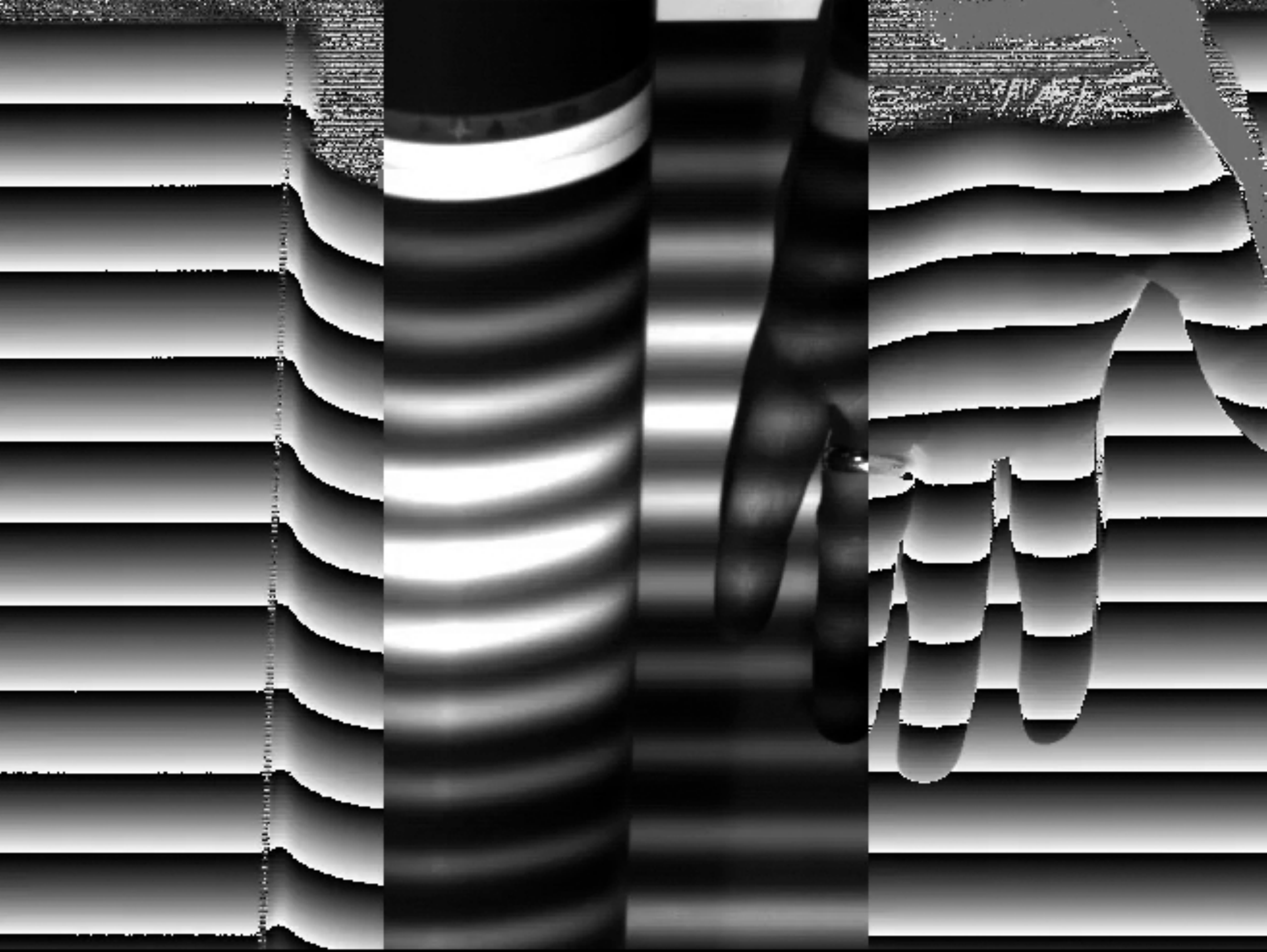}}
\caption{Phase reconstruction via dual-frequency PMP where the (left) and (right) thirds show the phase reconstruction as a pseudo-color image while (center) shows a frame of raw video.}
\label{fig00}
\end{figure}

Using the above procedure, Fig.~\ref{fig00} shows a frame from a video sequence demonstrating the use of PMP where the left and right thirds of the frame are pseudo-color plots the incoming video according to the projector row coordinates while the center section shows the raw video frame for a dual-frequency PMP sequence.    As a contour plot, the profile of the target surface becomes plainly evident. 

\section{Halftoning PMP Patterns}
Halftoning is the process of converting a continuous-tone image or photograph into binary black and white dots for display in digital printers, which can only choose to print or not print a dot. The illusion of continuous tone is then produced by the low-pass nature of the human visual system to blur the tiny dots at appropriate viewing distances from the printed page. In the case of using binary light modulators like the TI DLP chip, the projected image can, likewise, rely on defocusing the projector lens to produce the low-pass filtering otherwise expected of the human visual system.  As such, one would expect  traditional halftoning algorithms to produce the best binary images for mimicking contone PMP patterns.

\begin{figure}[t]
\centerline{\includegraphics[width=3.30in]{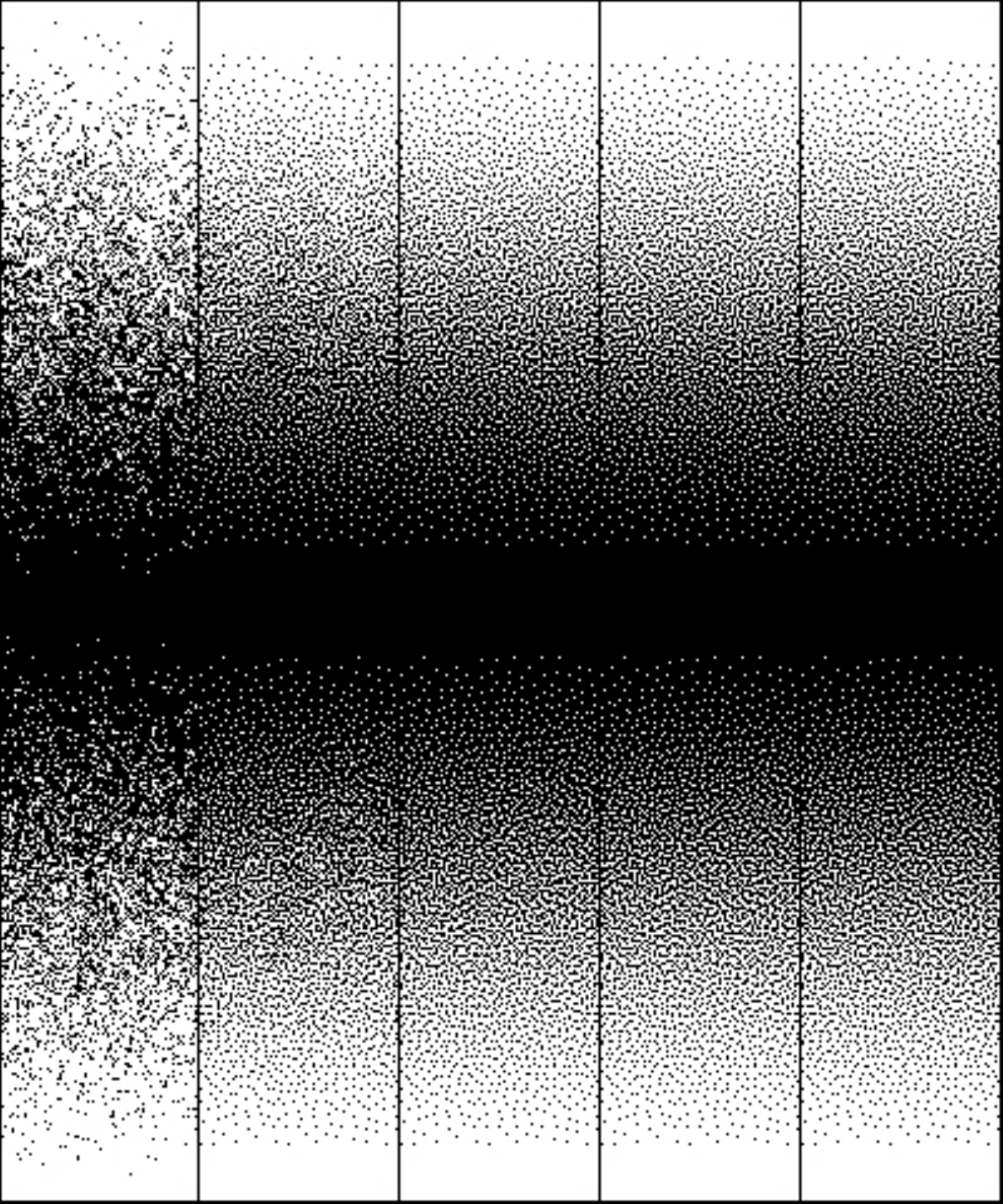}}
\caption{The progression of Direct Binary Search to construct a binary representation of a contone, unit-frequency, PMP pattern from a white-noise pattern to a fully optimized DBS pattern.}
\label{fig01}
\end{figure}

As stated previously, DBS is highly regarded for its ability to minimize low-frequency artifacts between the original and halftoned renditions. And in this process, a low-pass FIR filter kernel is used to model the human visual system.  Filtering the binary image with this filter then creates an estimate of the image as seen by a human observer.  Pixels are processed one-by-one starting in the top left and moving left-to-right and top-to-bottom. At each pixel, a decision is made to either toggle the pixel or swap it with one of its eight neighbors depending on which, if any, reduces the total error between the low-pass modeled, halftoned image and the target image being halftoned.  This process then repeats itself until a given level of quality is achieved. By replacing the human visual model with a model of the modulation transfer function of the defocused projector, DBS can produce binary renditions of PMP patterns as illustrated in Fig.~\ref{fig01}, which shows the progression of dots in the first of eight dual-frequency PMP patterns. 

As an attempt to minimize the phase error caused by halftoning, we propose a novel process that is also iterative and starts with initially halftoned pixels, $\tilde{I}_p[n]$, produced from the contone pixels, $I_p[n]$, by means of white-noise dithering where each of the $N$ values is randomly assigned a 0 or 1 with the likelihood of being assigned a 1 equal to the contone intensity value. Having halftoned the entire set of PMP patterns in this way, the next step is to process each pixel as a vector of $N$ binary digits, moving left-to-right and row-by-row.  After completing a pass through all pixels, the process then repeats until some error metric is appropriately minimized.  

During the processing of a given pixel, we pre-define a mean-preserving, low-pass, FIR filter modeling the MTF of the defocused projector, which in our case will be a $15\times 15$ Gaussian filter kernel with $\sigma=2.0$.  We then set the center pixel of this filter kernel equal to zero. By doing so, the filter produces a measure of how much light we can expect to blur into our target pixel from its neighbors, light that is separate from the pixel itself but contributing to its phase value all the same.  Using this measured light from neighboring pixels, our $15\times 15$ neighborhoods, from the $N$ patterns, become the scalar values of an $N$-length vector, which we define as $\tilde{I}_s[n]$ using $s$ to indicate light from the {\em surrounding} area.

\begin{figure}[t]
\centerline{\includegraphics[width=3.30in]{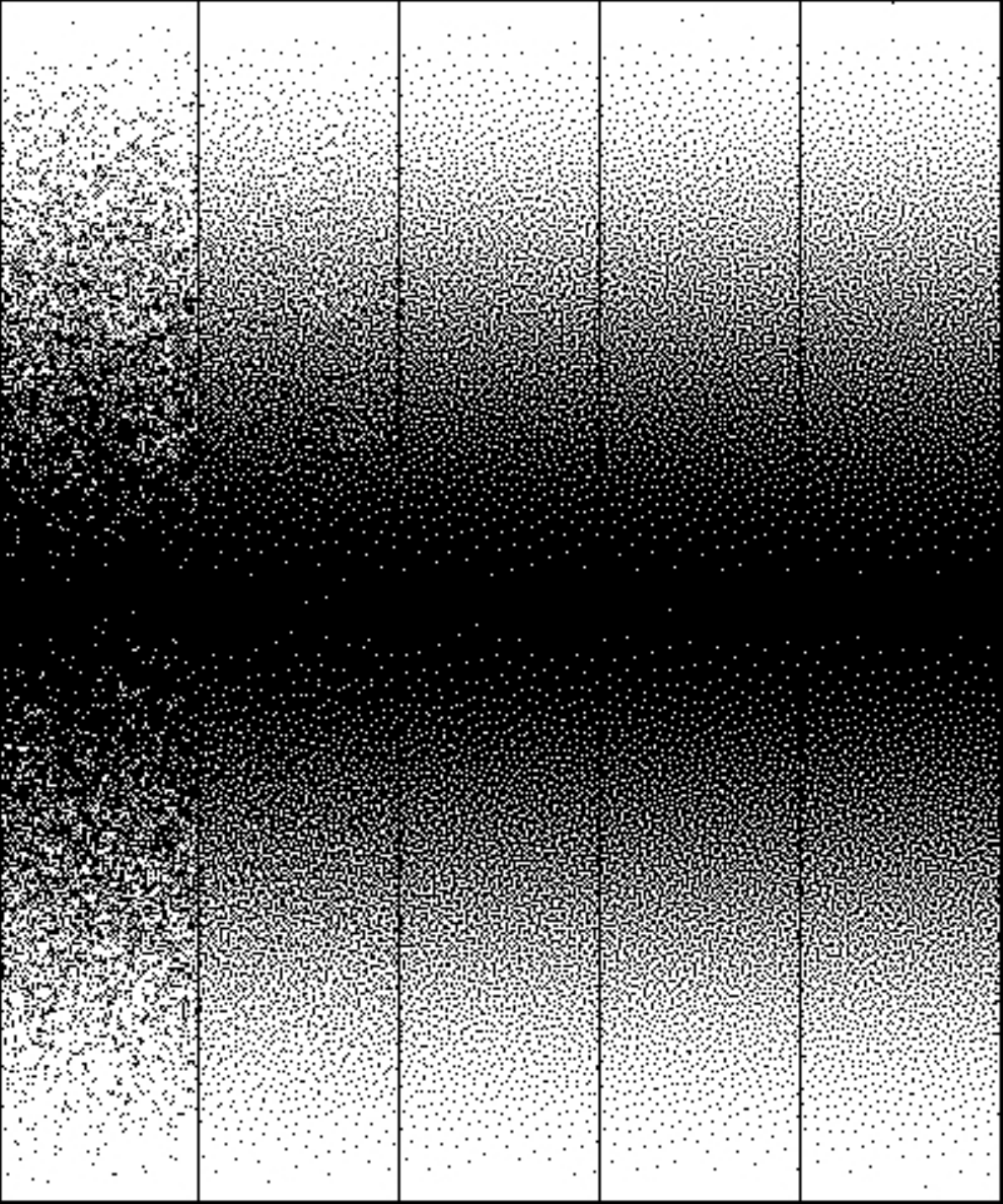}}
\caption{The progression of the proposed phase-weighted DBS, with $w_k=1$ for all $k$, to construct a binary representation of a contone, unit-frequency, PMP pattern from a white-noise pattern to a fully phase-optimized pattern.}
\label{fig02}
\end{figure}

\begin{figure}[t]
\centerline{\includegraphics[width=3.30in]{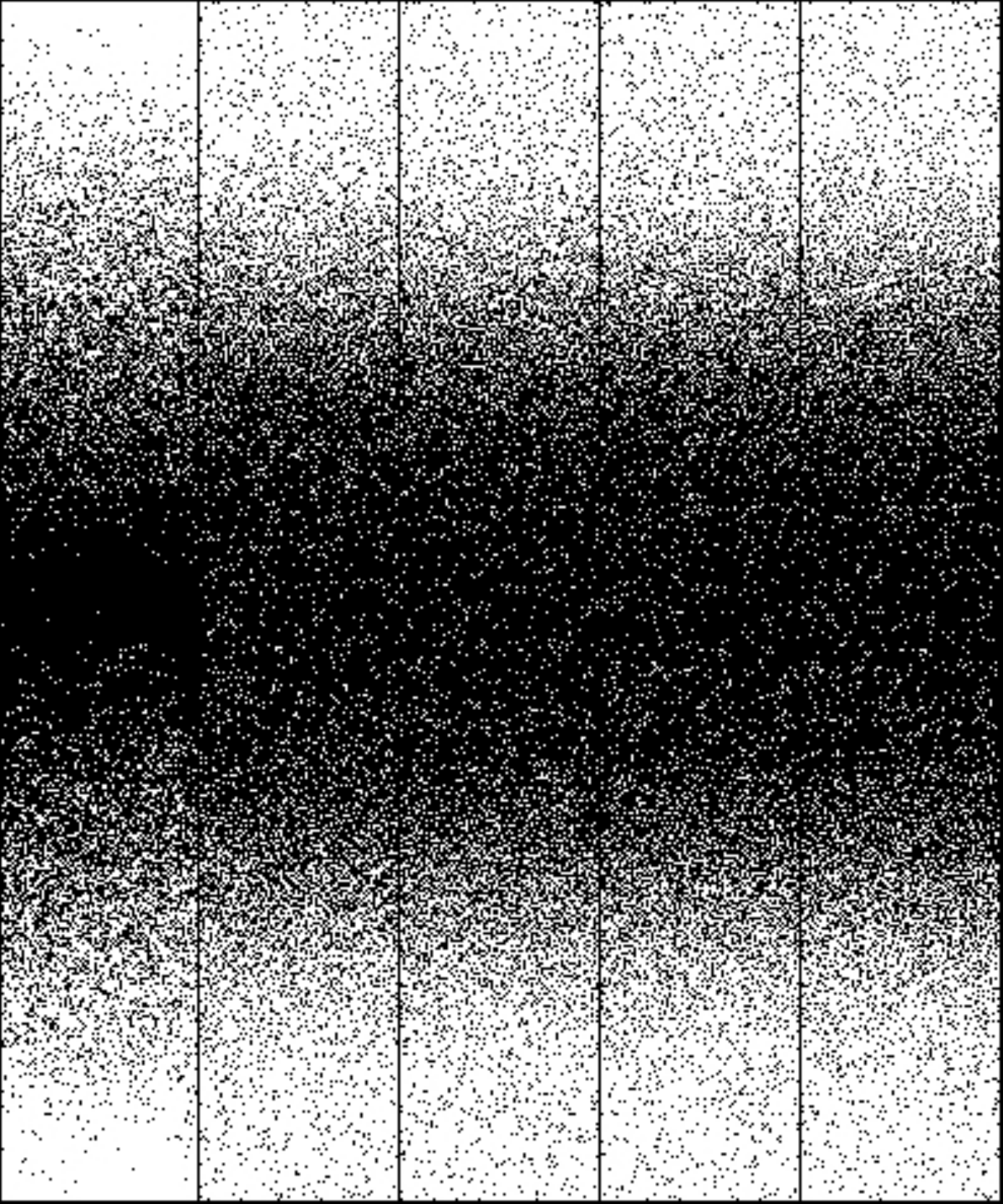}}
\caption{The progression of the proposed phase-weighted DBS, with $w_k=1$ for $k=1$ and $0$ otherwise, to construct a binary representation of a contone, unit-frequency, PMP pattern from a white-noise pattern to a fully phase-optimized pattern.}
\label{fig03}
\end{figure}

Having $\tilde{I}_s[n]$, we use an $N$-point DFT to produce $\tilde{I}_s[k]$, which we subtract from our ideal Fourier coefficients, $I_p[k]$ derived from the contone patterns, to produce the vector:
\begin{equation}
    I_d[k]  = I_p[k] - \tilde{I}_s[k],
\end{equation}
where $d$ indicates {\em difference}. This difference vector, $I_d[k]$, then represents the ideal DFT coefficients for our subject pixel sans halftoning.  The closest we can get to these DFT coefficients, from our halftoned vector, is then derived by identifying the best binary sequence, $I_b[n]$, whose DFT coefficients, $I_b[k]$, minimize the error:
\begin{equation}
\tilde{I}_p[n] = {\arg \min_{I_b[n]}} \sum_{k=0}^{N-1} w_k \| I_d[k] - c I_b[k] \|_2^2
\label{eqn::costFunction}
\end{equation}
where $c$ represents the weight of the center pixel from our low-pass, FIR filter modeling the projector MTF before we set it to 0. 

The weights, $w_k$, give us some freedom to allow certain DFT coefficients to have greater influence on the final result. For instance, we may weight all coefficients equally by setting $w_k = 1$ for all $k$ to create the optimized patterns in Fig.~\ref{fig02}.  From visual inspection, its clear that this approach is largely equivalent to traditional DBS and minimizing the spatial domain variance between the projected binary patterns and their contone counterparts.  Compared to the results of Fig.~\ref{fig01}, these new patterns show a slight amount of clustering most visible in the mid-tones. A far more fascinating result occurs when we focus our algorithm at minimizing phase error on the $k=1$ coefficient where we set $w_k = 1$ for $k=1$ and 0, otherwise, to create the optimized patterns in Fig.~\ref{fig03}.  From visual inspection, these patterns look awful; however, its not for visual inspection that these patterns are built.  Instead, they are tuned to add grain in space in order to minimize phase error in time.  

Of course for large values of $N$, the total search space for finding the best bit combination might be so large as to make exhaustive search impractical. So a very simple approximation of eqn.~(\ref{eqn::costFunction}) can be performed by applying the inverse DFT to $I_d[k]$ to produce $I_d[n]$, and then thresholding the resulting contone values according to:
\begin{equation}
\tilde{I}_p[n] = \left\{\begin{array}{rl}
                   1, & \mbox{if } I_d[n] > 0\\
                   0, & \mbox{else}
                  \end{array}. \right.
\label{eqn::bestDftCoefficients}
\end{equation}
Regardless, the processing of the current pixel is now complete and moves to the next pixel along the raster path.  

\section{Experimental Results}
In order to evaluate the theoretical improvement in producing phase images using the new binary pattern construction algorithm, we will consider two pattern schemes.  The first will be traditional PMP as defined by eqn.~(\ref{EQ:CameraPattern}), and second will be Liu {\it et al}'s dual-frequency pattern scheme defined by:
\begin{eqnarray}
I_p[n] & = & \frac{1}{2} + \frac{1}{4} \cos \left( 2\pi (\frac{n}{N}-y_p)\right) + \ldots \nonumber \\
	& &  \ldots + \frac{1}{4} \cos \left(4\pi (\frac{n}{N}-8y_p)\right).
\label{EQ:ProjectorPattern}
\end{eqnarray}
Regardless of the parameters used in eqn.~(\ref{eqn::costFunction}), the quality of the final pattern set will be defined by the total sum of absolute errors in the phase of the $k=1$ coefficient, for single frequency PMP, and the $k=2$ coefficient, for dual. Direct binary search will be used as the baseline in either case. White noise dither patterns will be used as the seed patterns in all cases.

\begin{figure}[t]
\centerline{\includegraphics[width=3.30in]{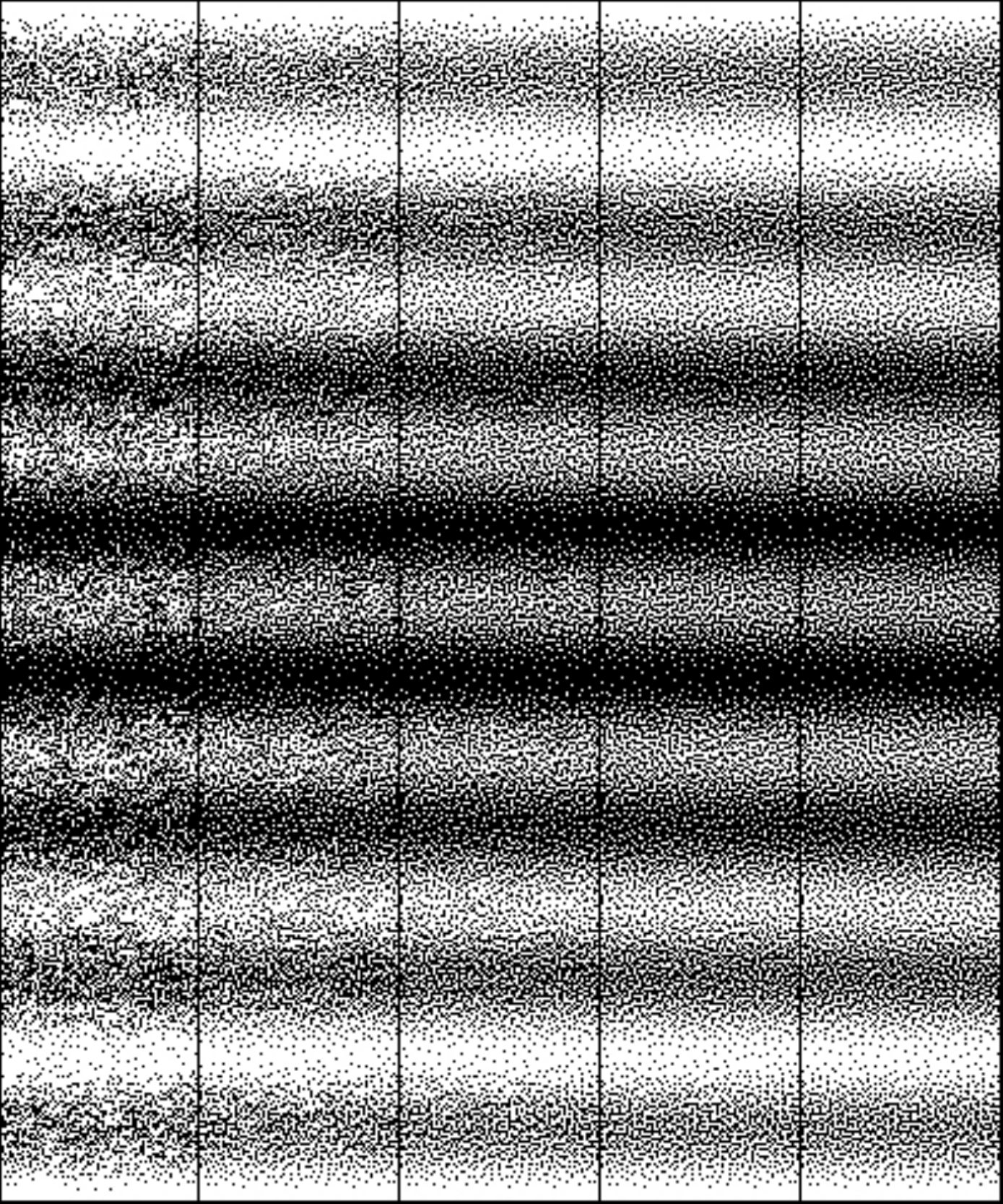}}
\caption{The progression of the proposed phase-weighted DBS, with $w_k=1$ for all $k$, to construct a binary representation of a contone, dual-frequency, PMP pattern from a white-noise pattern to a fully phase-optimized pattern.}
\label{fig04}
\end{figure}

\begin{figure}[t]
\centerline{\includegraphics[width=3.30in]{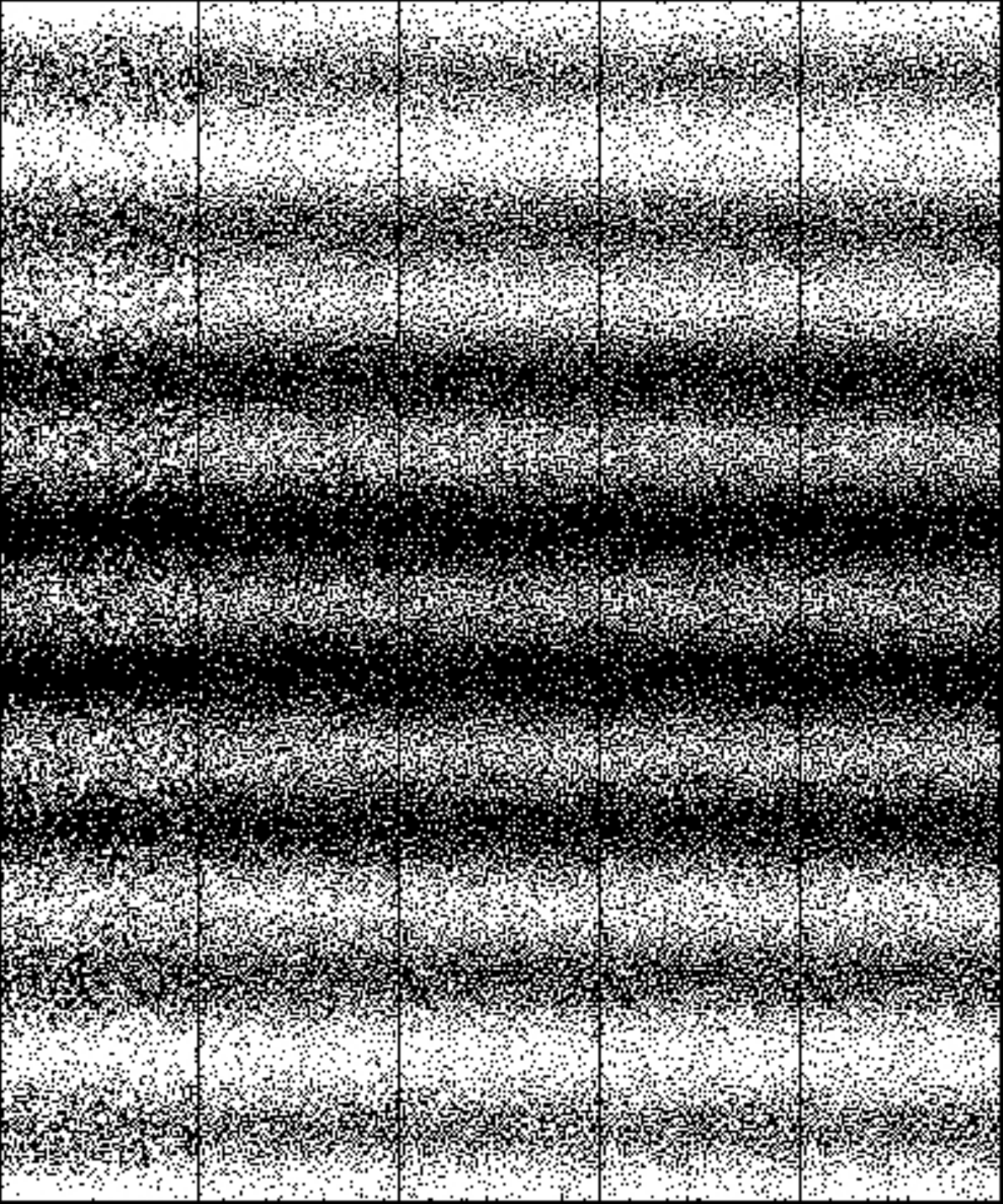}}
\caption{The progression of the proposed phase-weighted DBS, with $w_k=1$ for $k=1$ and $0$ otherwise, to construct a binary representation of a contone, dual-frequency, PMP pattern from a white-noise pattern to a fully phase-optimized pattern.}
\label{fig05}
\end{figure}

\begin{figure}[t]
\centerline{\includegraphics[width=3.30in]{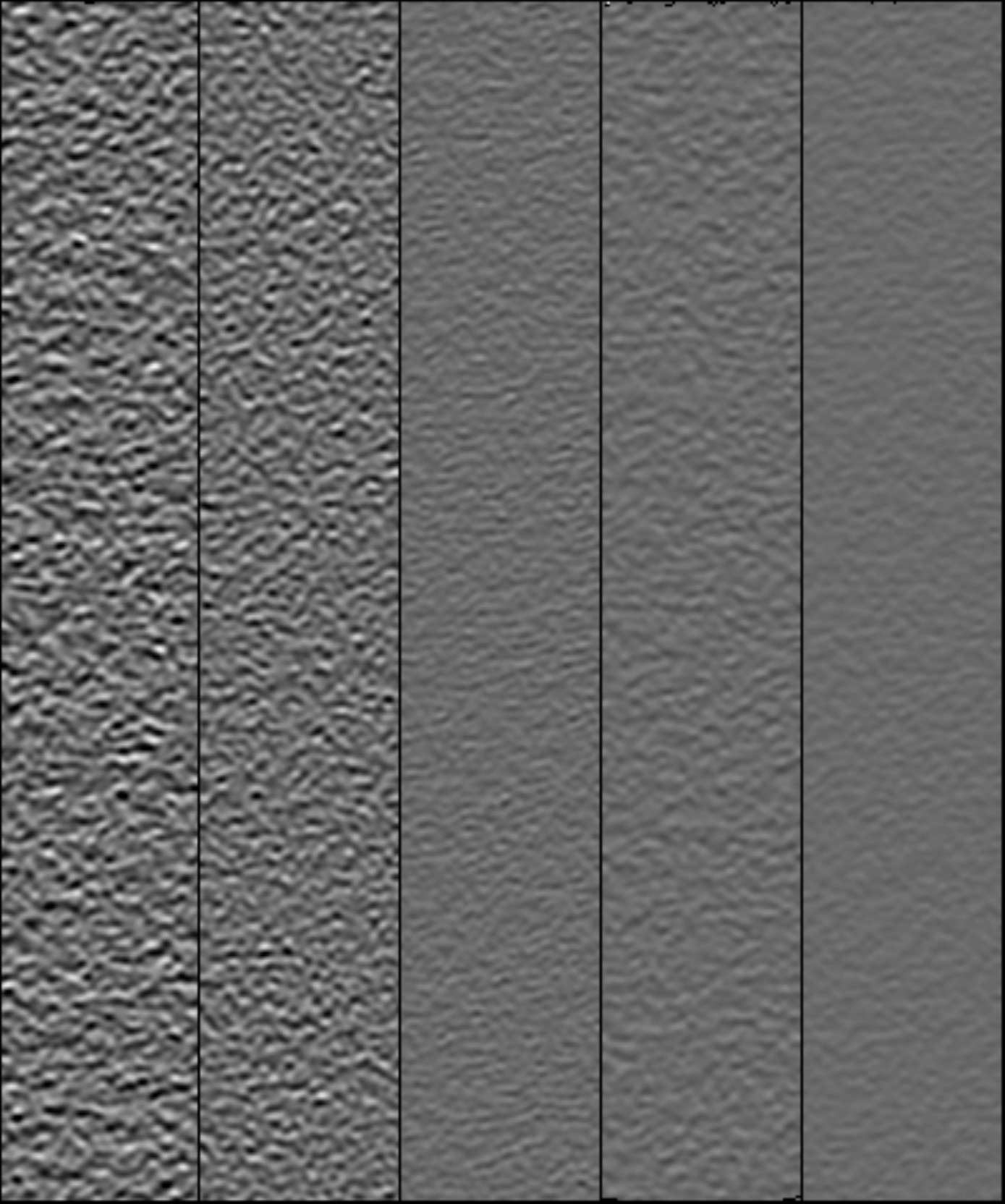}}
\caption{Surface gradients produced by the PMP patterns of Figs.~\ref{fig01}-\ref{fig06} from left to right, respectively.}
\label{fig06}
\end{figure}

To see the new dithering procedure in action, Figs.~\ref{fig02} and \ref{fig03} show the evolution in the $1^{st}$ of 8 patterns for single frequency PMP patterns while Figs.~\ref{fig04} and \ref{fig05} show the corresponding evolution for dual frequency.  Shown in Fig.~\ref{fig06} are surface gradients produced by the patterns of Figs.~\ref{fig01} through \ref{fig05} moving left-to-right, respectively. As will be the case in all examples, images will be constructed as $80\times 480$ images with left-to-right wrap around so that images can be tiled side-to-side to form $640 \times 480$ images for VGA projectors. Error in the patterns will be calculated as the mean absolute error in degrees. In the case of Figs.~\ref{fig01}-\ref{fig03}, patterns represent single frequency PMP where phase values range from 0 to 360 degrees from top to bottom.  For the dual frequency patterns of Figs.~\ref{fig04}-\ref{fig05}, the phase error is determined prior to unwrapping where phase values range from 0 to 2,880 degrees.

In the particular example of Fig.~\ref{fig02}, the patterns were optimized over the 16 iterations with all DFT coefficients evaluated equally in eqn.~(\ref{eqn::costFunction}). As listed in Table~\ref{tab01} where we assume an ideal projector MTF, the mean absolute error in phase drops from $2.79$ to $0.43$ degrees per pixel, which is not quite as good as the DBS pattern in Fig.~\ref{fig01} with a mean absolute phase error of $0.43$ degrees, but the new technique does perform better in Fig.~\ref{fig03} when we iterate on the $k=1$ coefficient exclusively where 28 iterations results in a mean absolute phase error of $0.10$ degrees per pixel, a $3\times$ improvement over DBS.

Visually, the patterns of Fig.~\ref{fig03} are interesting because no reasonable halftoning algorithm would produce patterns of such poor visual quality and only through the correlation of patterns could one consider them better than those produced by DBS. This difference in clearly visible in Fig.~\ref{fig04} where we show the gradient on the phase images. The reason for the improvement is quite obvious when we calculate the average power in the DFT coefficients of the $I_d[k]$ images where the $I_d[1]$ image has an average power of $3.38e-5$ while all others range in value from $1.67e-2$ ($k=0$) to $7.95e-2$ ($k=3$). The new algorithm is simply moving the dithering noise from $I_d[1]$ to the others coefficients, whereas for $w_k=1$ for all $k$, the new algorithm has to spread dither noise across all coefficients equally at approximately $6.79e-4$ in Fig.~\ref{fig02}. 

\begin{table}
\caption{Theoretical phase errors comparing spatial DBS with the proposed phase DBS where phase is calculated with a simulated projector MTF.}
\begin{center}
    \begin{tabular}{|l|c|}
    \hline
    \multicolumn{2}{|c|}{Unit Frequency} \\
    \hline
Optimization Scheme & Mean Abs. \\ 
     & Error (deg) \\ \hline
spatial DBS &  0.43 \\
phase DBS ($w_k=1$ for all $k$) &  0.44 \\
phase DBS ($w_k=1$ for $k=1$ else 0) &  0.10 \\
\hline
\multicolumn{2}{|c|}{Dual Frequency} \\
\hline
spatial DBS &  0.75 \\
phase DBS ($w_k=1$ for all $k$) &  0.87 \\
phase DBS ($w_k=1$ for $k=1,2$ else 0) &  0.44 \\
    \hline
    \end{tabular}
\end{center}
\label{tab01}
\end{table}

\begin{table}[b]
\caption{Experimental phase errors comparing spatial DBS with the proposed phase DBS using a SLI scanner composed of a VGA projector and Prosilica GC640 machine vision camera.}
\begin{center}
    \begin{tabular}{|l|c|}
    \hline
    \multicolumn{2}{|c|}{Unit Frequency} \\
    \hline
Optimization Scheme & Mean Abs. \\ 
     & Error (deg) \\ \hline
spatial DBS &  $6.34e-3$ \\
phase DBS ($w_k=1$ for $k=1$ else 0) &  $3.47e-3$ \\
\hline
\multicolumn{2}{|c|}{Dual Frequency} \\
\hline
spatial DBS &  $3.43e-2$ \\
phase DBS ($w_k=1$ for $k=1,2$ else 0) &  $2.36e-2$ \\
    \hline
    \end{tabular}
\end{center}
\label{tab02}
\end{table}

As a comparison to prior works, we note that, in Dai~{\it et al}~\cite{Dai201479}, the author's simulated  projector defocus by using a Gaussian filter of size $5\times 5$ with a variance of $5/3$, which they argued modeled a projector just-out-of-focus. They then ran their proposed genetic algorithm and showed that for a single period of 18 pixels, they reduced the phase noise for a Bayer's dither PMP pattern from a root-mean-squared error of 0.064 down to approximately 0.031 radians.  Their algorithm was limited to using a single, spatially shifted pattern with three steps.  So its not a direct comparison; however, applying the same Gaussian filter on our 8 unique patterns with a sinusoidal period of 32 pixels, we measured an RMS error of 0.047 radians for a Bayer dithered pattern set, which was reduced to 0.027 radians via DBS and 0.014 via the proposed phase DBS algoritm.  For comparison, Dai~{\it et al} reported an improved down to 48\% of the Bayer's dither error while we are reporting 57\% for DBS and 29\% for phase DBS.

Looking now at dual-frequency PMP, Fig.~\ref{fig04} shows the binary patterns after 14 iterations where the mean absolute error in phase ($k=2$) drops from $2.79$ to $0.87$ degrees per pixel when weighing all DFT coefficients equally. In contrast, DBS (not shown) achieves a phase error of $0.75$ degrees per pixel. If we now only include the $k=1$ and $k=2$ coefficients in eqn.~(\ref{eqn::costFunction}), then the phase error drops to $0.44$ degrees per pixel in Fig.~\ref{fig05}. Again from visual inspection, the resulting patterns are far from ideal in terms of traditional halftoning metrics; yet, they still score a $40\%$ reduction in noise over DBS in terms of phase error. 

\begin{figure}[t]
	\centerline{\includegraphics[width=3.30in]{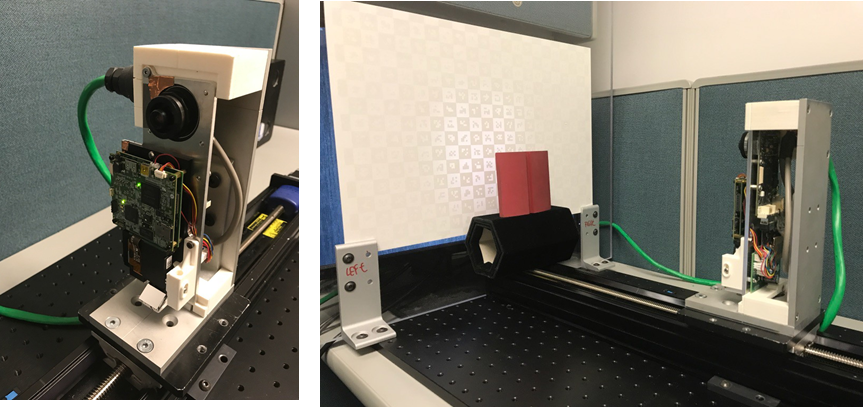}}
	\caption{Our scanner with a split screen pattern with one side having a traditional DBS pattern set and the second side the proposed phase optimized DBS algorithm.}
	\label{setup}
\end{figure}
\begin{figure}[t]
	\centerline{\includegraphics[width=3.30in]{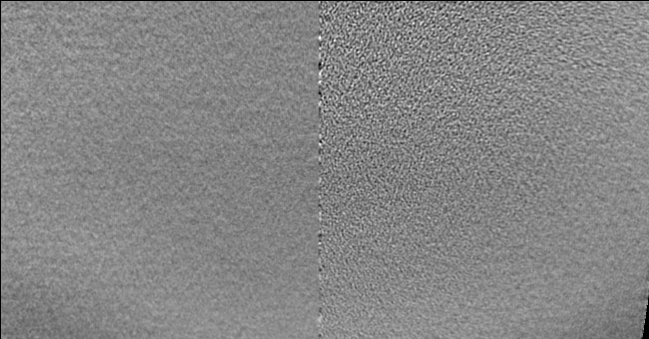}}
	\caption{The reconstruction of the flat background.}
	\label{flat}
\end{figure}

In order to evaluate our proposed halftoning process in a physical device, we assembled a scanner composed of a $640\times 480$ DLP projector running at 60 frames per second where patterns were delivered to the projector over VGA using fragment shaders on the host PC's graphics card. Capturing images was a $659\times 494$ pixel Prosilica GC640, synchronized to the projector by means of the VGA signal's vertical sync. The camera was set to use a 4 millisecond exposure while the camera lens iris was adjusted to avoid over/under saturation of the sensor. Defocusing of the projector was performed from visual inspection of the projected image on a white foam board. Thermal noise was removed by averaging 10 separate scans together to form a single phase image. As shown in Fig.~\ref{setup}, we loaded the scanner with a split screen pattern with one side having a traditional DBS pattern set and the second side the proposed phase optimized DBS algorithm. The pattern set is the single frequency set from Fig. 3 in the paper. Note that the projector is aimed down with a mirror used to direct the light forward, creating a keystone distortion on the projected patterns. Shown in Fig.~\ref{flat} is the reconstruction of the flat background. Fitting a plane to each half, the phase DBS algorithm (left) has a mean absolute error of 26.6 microns while the traditional DBS corresponds to 43.2 microns, which is a reduction in error of 38.43\%. Table~\ref{tab02} summarized the measured phase errors for these live video tests.

Using the single-frequency patterns of Fig.~\ref{fig02} which were optimized solely on the $k=1$ DFT coefficient, the resulting variance in the phase error, from camera video, measured $3.47e-3$ degrees per pixel while DBS scored $6.34e-3$ degrees per pixel, an approximately $45\%$ reduction in halftone texture. Using the dual-frequency patterns of Fig.~\ref{fig05} optimized over the $k=1$ and $k=2$ coefficients, the resulting phase error variance, from camera video, measured $2.36e-2$ degrees per pixel while DBS scored $3.43e-2$ degrees per pixel, a reduction of approximately $30\%$.

\section{Conclusions}
In this paper, we introduced a novel procedure for building binary dither patterns that mimic contone patterns associated with phase-measuring profilometry. The advantage to using these patterns is that, in binary light modulators, these patterns become immune to the effects of short camera exposures and, as such, offer greater flexibility for scanning at high speeds. As demonstrated both theoretically and experimentally, the new procedure proved itself at minimizing spatial artifacts in a wide range of PMP coding schemes, having produced patterns for dual frequency PMP as well as for detecting motion. What is particularly interesting in these results is the visual difference between patterns produced by means of DBS and of the new algorithm. In terms of traditional halftoning metrics, the patterns produced by the new algorithm exhibit excessive noise/grain at all gray-levels in the pattern that a human observer would find objectionable and that DBS eliminates.


{\small
\bibliographystyle{ieee}
\bibliography{egbib}

\begin{thebibliography}{10}\itemsep=-1pt

\bibitem{REF:AnalouiAllebachDirectBinarySearch}
M.~Analoui and J.~P. Allebach.
\newblock Model-based halftoning using direct binary search.
\newblock In {\em Proc. SPIE}, volume 1666, pages 96--108, 1992.

\bibitem{REF:Chen2013SLM}
J.~Chen, X.~Lv, and S.~Zeng.
\newblock Doubling the resolution of spatial-light-modulator-based differential
  interference contrast microscopy by structured illumination.
\newblock {\em Opt. Lett.}, 38(17):3219--3222, Sep 2013.

\bibitem{REF:Dai2013DitherPattern}
J.~Dai, B.~Li, and S.~Zhang.
\newblock High-quality fringe pattern generation using binary pattern
  optimization through symmetry and periodicity.
\newblock {\em Optics and Lasers in Engineering}, 52:195 -- 200, 2014.

\bibitem{Dai201479}
J.~Dai, B.~Li, and S.~Zhang.
\newblock Intensity-optimized dithering technique for three-dimensional shape
  measurement with projector defocusing.
\newblock {\em Optics and Lasers in Engineering}, 53:79 -- 85, 2014.

\bibitem{Dai2013790}
J.~Dai and S.~Zhang.
\newblock Phase-optimized dithering technique for high-quality 3d shape
  measurement.
\newblock {\em Optics and Lasers in Engineering}, 51(6):790 -- 795, 2013.

\bibitem{REF:Geng2011SLITutorial}
J.~Geng.
\newblock Structured-light 3d surface imaging: a tutorial.
\newblock {\em Adv. Opt. Photon.}, 3(2):128--160, Jun 2011.

\bibitem{REF:Gong2010SLIUltraFast}
Y.~Gong and S.~Zhang.
\newblock Ultrafast 3-d shape measurement with an off-the-shelf dlp projector.
\newblock {\em Opt. Express}, 18(19):19743--19754, Sep 2010.

\bibitem{REF:Kawasaki2013OneShot}
H.~Kawasaki, H.~Masuyama, R.~Sagawa, and R.~Furukawa.
\newblock Single colour one-shot scan using modified penrose tiling pattern.
\newblock {\em IET Computer Vision}, 7(5):293--301, October 2013.

\bibitem{REF:Khoshelham2012Kinect}
K.~Khoshelham and S.~O. Elberink.
\newblock Accuracy and resolution of kinect depth data for indoor mapping
  applications.
\newblock {\em Sensors}, 12(2):1437, 2012.

\bibitem{REF:Lau2010MotionDetection}
D.~L. Lau, K.~Liu, and L.~G. Hassebrook.
\newblock Real-time three-dimensional shape measurement of moving objects
  without edge errors by time-synchronized structured illumination.
\newblock {\em Opt. Lett.}, 35(14):2487--2489, Jul 2010.

\bibitem{Lei:09}
S.~Lei and S.~Zhang.
\newblock Flexible 3-d shape measurement using projector defocusing.
\newblock {\em Opt. Lett.}, 34(20):3080--3082, Oct 2009.

\bibitem{Li2014236}
B.~Li, Y.~Wang, J.~Dai, W.~Lohry, and S.~Zhang.
\newblock Some recent advances on superfast 3d shape measurement with digital
  binary defocusing techniques.
\newblock {\em Optics and Lasers in Engineering}, 54:236 -- 246, 2014.

\bibitem{REF:Liu2010Realtime3D}
K.~Liu, Y.~Wang, D.~L. Lau, Q.~Hao, and L.~G. Hassebrook.
\newblock Dual-frequency pattern scheme for high-speed 3-d shape measurement.
\newblock {\em Opt. Express}, 18(5):5229--5244, Mar 2010.

\bibitem{REF:Liu2011PhaseModulation}
Y.~Liu, X.~Su, and Q.~Zhang.
\newblock A novel encoded-phase technique for phase measuring profilometry.
\newblock {\em Opt. Express}, 19(15):14137--14144, Jul 2011.

\bibitem{Lohry2012917}
W.~Lohry and S.~Zhang.
\newblock 3d shape measurement with 2d area modulated binary patterns.
\newblock {\em Optics and Lasers in Engineering}, 50(7):917 -- 921, 2012.

\bibitem{REF:Mosino2010PSIQuadrature}
J.~F.~M. {n}o, J.~C. Guti\'{e}rrez-Garc\'{i}a, T.~A. Guti\'{e}rrez-Garc\'{i}a,
  and J.~M. Mac\'{i}as-Preza.
\newblock Two-frame algorithm to design quadrature filters in phase shifting
  interferometry.
\newblock {\em Opt. Express}, 18(24):24405--24411, Nov 2010.

\bibitem{REF:Salvi2010SLIReview}
J.~Salvi, S.~Fernandez, T.~Pribanic, and X.~Llado.
\newblock A state of the art in structured light patterns for surface
  profilometry.
\newblock {\em Pattern Recognition}, 43(8):2666 -- 2680, 2010.

\bibitem{REF:Salvi2004SLIReview}
J.~Salvi, J.~Pag{\`e}s, and J.~Batlle.
\newblock Pattern codification strategies in structured light systems.
\newblock {\em Pattern Recognition}, 37(4):827 -- 849, 2004.
\newblock Agent Based Computer Vision.

\bibitem{REF:Su2007ColorSLI}
W.-H. Su.
\newblock Color-encoded fringe projection for 3d shape measurements.
\newblock {\em Opt. Express}, 15(20):13167--13181, Oct 2007.

\bibitem{REF:Sun2013Imaging3D}
B.~Sun, M.~Edgar, R.~Bowman, L.~Vittert, S.~Welsh, A.~Bowman, and M.~Padgett.
\newblock 3-dimensional computational ghost imaging.
\newblock In {\em Frontiers in Optics 2013}, page FW5D.2. Optical Society of
  America, 2013.

\bibitem{Sun2015158}
J.~Sun, C.~Zuo, S.~Feng, S.~Yu, Y.~Zhang, and Q.~Chen.
\newblock Improved intensity-optimized dithering technique for 3d shape
  measurement.
\newblock {\em Optics and Lasers in Engineering}, 66:158 -- 164, 2015.

\bibitem{ulichney}
R.~A. Ulichney.
\newblock Dithering with blue noise.
\newblock {\em Proceedings of the IEEE}, 76(1):56--79, Jan. 1988.

\bibitem{REF:Wang2011SLICoding}
Y.~Wang, K.~Liu, Q.~Hao, D.~L. Lau, and L.~G. Hassebrook.
\newblock Period coded phase shifting strategy for real-time 3-d structured
  light illumination.
\newblock {\em IEEE Transactions on Image Processing}, 20(11):3001--3013, Nov
  2011.

\bibitem{REF:Wissmann2011SLM}
P.~Wissmann, F.~Forster, and R.~Schmitt.
\newblock Fast and low-cost structured light pattern sequence projection.
\newblock {\em Opt. Express}, 19(24):24657--24671, Nov 2011.

\bibitem{doi:10.1117/1.2402128}
S.~Zhang and P.~S. Huang.
\newblock High-resolution, real-time three-dimensional shape measurement.
\newblock {\em Optical Engineering}, 45(12):123601--123601--8, 2006.

\bibitem{REF:Zhang2010SLISuperFast}
S.~Zhang, D.~V.~D. Weide, and J.~Oliver.
\newblock Superfast phase-shifting method for 3-d shape measurement.
\newblock {\em Opt. Express}, 18(9):9684--9689, Apr 2010.

\bibitem{Zuo:12}
C.~Zuo, Q.~Chen, S.~Feng, F.~Feng, G.~Gu, and X.~Sui.
\newblock Optimized pulse width modulation pattern strategy for
  three-dimensional profilometry with projector defocusing.
\newblock {\em Appl. Opt.}, 51(19):4477--4490, Jul 2012.

\end{thebibliography}
}

\end{document}